\documentclass[runningheads]{llncs}
\usepackage{amsmath,amssymb,amsfonts}
\usepackage{graphicx}
\usepackage{xcolor}
\usepackage{colortbl}
\usepackage{hhline}
\usepackage{booktabs}
\usepackage[numbers,sectionbib]{natbib}
\usepackage{pgfplots}
\usepackage{subcaption}
\usepackage{multirow}

\newcommand{\com}[1]{{\leavevmode\color{blue}COM: #1}}
\newcommand{\res}[1]{{\leavevmode\color{green}RES: #1}}
\newcommand{\tod}[1]{{\leavevmode\color{red}TOD: #1}}
\newcommand{\raus}[1]{{\leavevmode\color{gray}RAUS: #1}}
\newcommand{\drin}[1]{#1}

\newcommand{\umschreiben}[1]{{\leavevmode\color{green}Umschreiben: #1}}

\usepackage{enumitem}
\setitemize{topsep=0.1em,parsep=0.1em,partopsep=0.1em}

\allowdisplaybreaks[4]
\renewcommand{\com}[1]{}
\renewcommand{\res}[1]{}
\renewcommand{\tod}[1]{}

\renewcommand{\raus}[1]{}
\renewcommand{\umschreiben}[1]{#1}

\newcommand{\vektor}[1]{\ensuremath{\mathrm{\mathbf{#1}}}} 
\newcommand{\vx}{\vektor{x}}
\newcommand{\vy}{\vektor{y}}
\newcommand{\vp}{\vektor{p}}

\newcommand{\ds}[1]{\textsc{#1}}
\newcommand{\ap}[1]{\textsl{#1}}

\begin{document}
%
\title{Extreme Gradient Boosted Multi-label Trees for Dynamic Classifier Chains}
%
%
\author{Simon Bohlender \and
Eneldo Loza Menc\'ia \and Moritz Kulessa}
\authorrunning{S. Bohlender, E. Loza Menc\'ia, M. Kulessa}
%
\institute{Knowledge Engineering Group,  Technische Universit\"at Darmstadt, \\Hochschulstr. 10, 
64289 Darmstadt, Germany \\
\email{simon.bohlender@gmail.com}, \email{\{eneldo,mkulessa\}@ke.tu-darmstadt.de}
}
\maketitle              
\begin{abstract}
Classifier chains is a key technique in multi-label classification, since it allows to consider label dependencies effectively. 
However, the classifiers are aligned according to a static order of the labels. 
In the concept of dynamic classifier chains (DCC) the label ordering is chosen for each prediction dynamically depending on the respective instance at hand. 
We combine this concept with the boosting of extreme gradient boosted trees (XGBoost), an effective and scalable state-of-the-art technique, and incorporate DCC in a fast multi-label extension of XGBoost which we make publicly available. 
As only positive labels have to be predicted and these are usually only few, the training costs can be further substantially reduced. 
Moreover, as experiments on eleven datasets show, the length of the chain allows for a more control over the usage of previous predictions and hence over the measure one want to optimize.
\keywords{multi-label classification \and classifier chains \and gradient boosted trees}
\end{abstract}
\section{Introduction\com{(max 1 page. shorten also abstract?)}}
\tod{include related work:
https://arxiv.org/pdf/1909.04373.pdf

[1] Si S, Zhang H, Keerthi S S, et al. Gradient boosted decision trees 
for high dimensional sparse output[C]//Proceedings of the 34th 
International Conference on Machine Learning-Volume 70. JMLR. org, 2017: 
3182-3190. https://dl.acm.org/doi/10.5555/3305890.3306010 http://proceedings.mlr.press/v70/si17a.html
[2] Haobo Wang, Weiwei Liu, Yang Zhao, Chen Zhang, Tianlei Hu, and Gang 
Chen. Discriminative and correlative partial multi-label learning. In 
IJCAI, pages 3691–3697, 2019.
[3] Yan R, Tesic J, Smith J R. Model-shared subspace boosting for 
multi-label classification[C]//Proceedings of the 13th ACM SIGKDD 
international conference on Knowledge discovery and data mining. 2007: 
834-843.
[4] Wu F, Han Y, Tian Q, et al. Multi-label boosting for image 
annotation by structural grouping sparsity[C]//Proceedings of the 18th 
ACM international conference on Multimedia. 2010: 15-24.
}
Classical supervised learning tasks deal with the problem to assign a single class label to an instance. Multi-label classification (MLC) is an extension of these problems where each instance can be associated with multiple labels from a given label space \citep{MultiLabel-Overview}. 
A straight-forward solution, referred to as binary relevance decomposition (BR), learns a separate classification model for each of the target labels.
However, it neglects possible interactions between the labels. 
Classifier chains (CC) similarly learn one model per label, but these take the predictions of the previous models along a predetermined sequence of the labels into account \citep{cc}. 
It was shown formally that CC is able to capture local as well as global dependencies and that these are crucial if the goal is to predict the correct label combinations, rather than each label for itself \citep{dembczynski12PCCdependence}. 
However, in practice the success of applying CC highly depends on the order of the labels along the chain. 
Finding a good sequence is a non-trivial task.
\com{Hier 1-2 Sätze wegnehmen?}
First, the number of possible sequences to consider grows exponentially with the number of labels.
Second, even though a sequence might exist which is optimal w.r.t. some global dependencies in the data, local dependencies make it necessary to consider different chains for different instances.
For instance, in a driving scene scenario it is arguable easier to detect first a car and then infer its headlights during the day, whereas it is easier to first detect the lights and from that deduce the car during the night.\com{das beispiel ist nicht ganz schoen beschrieben}
Roughly speaking, each instance has its own sequence of best inferring its true labels. 

\emph{Dynamic chain} approaches address the problem of finding a good sequence for a particular instance. 
For instance, \citet{mk:DS-18-DCC-RDT} propose to build an ensemble of random decision trees (RDT) with special label tests at the inner nodes. 
The approach predicts at each iteration the label for which the RDT is most certain and re-uses that information in subsequent iterations. 
Despite the appealing simplicity due to the flexibility of RDT, it comes at the expense of predictive performance since RDT 
are not trained in order to optimize a particular measure.

The Extreme Gradient Boosted Trees \emph{(XGBoost)} approach 
\citep{chen2016xgboost}, 
instead, is a highly optimized and efficient tree induction method which has been very successful recently in international competitions.\tod{find good refs} 
Similarly to CC and dynamic chain approaches, XGBoost refines its predictions in subsequent iterations by using boosting. 
This served as inspiration to the proposed approach \emph{XDCC},\footnote{Publicly available at \url{https://github.com/keelm/XDCC}} which integrates \emph{\textbf{D}ynamic \textbf{C}lassifier \textbf{C}hains} into the e\textbf{x}treme boosting structure of gradient boosted trees.
XDCC's optimization goal in each boosting round is to predict only a single label for which it is the most certain.
This label can be different for each training instance and depends on the given data, label dependencies and previous predictions for the instance at hand.
The information about the predicted labels is carried over to subsequent rounds. 

A key advantage of the proposed approach is the reduced run time in comparison to classifier chains. 
This is due to the fact that though the total number of labels can be quite high in MLC, the number of actually relevant labels for each instance is relatively low, usually below 10. 
Hence, only few rounds are potentially enough if only the positive labels are predicted, whereas CC-based approaches have to still make predictions for each of the existing labels. 

\raus{
Our contributions are the following:\com{could be just put in the abstract}
 \begin{itemize}
     \item 
     We introduce a multi-label formulation of the XGBoost objective which is much more efficient than the decomposition based XGBoost baselines. The implementation is made publicly available.
      \item 
     We propose a variant of XGBoost which integrates dynamic classifier chains. In contrast to its static counterpart CC, it allows to trade-off between computational costs and the usage of previous predictions. Hence, it allows a more fine-grained control over the exploitation of label dependencies.
  \end{itemize}
}

\section{Preliminaries\com{(1-1,5 pages)}}
This section provides a short overview of the notations used in this work. Additionally an insight to XGBoosts basic functionality, i.e. to the tree boosting and classification process as well as the way it can deal with multiple targets, is given.
\subsection{Multilabel classification}
Multilabel classification (MLC) is the task of predicting for a finite set of $N$ unique class labels $\mathcal{\Lambda}=\{\lambda_1,\ldots,\lambda_N\}$ whether they are relevant/positive, i.e., $y_j=1$ if $\lambda_j$ is relevant, or $y_j=0$ if $\lambda_j$ is irrelevant/negative, for a given instance. 
The training set consists of training examples $\vx_i \in \mathcal{X}$ and associated label sets $\vy_i \in \mathcal{Y}=\{0,1\}^N$,  $1 \le i \le M$, which can be represented as matrices $X=(x_{ik}) \in A^{M\times K}$ and $Y=(y_{ij})\in \{0,1\}^{M\times N}$, where  features $x_{ij}$ can be represented as continuous, categorical or binary values.
An MLC classifier $h: \mathcal{X} \rightarrow \mathcal{Y}$ is trained on the training set in order to learn the mapping between input features and output label vector. The prediction of $h$ for a test example $\vx$ is a binary vector $\hat\vy=h(\vx)$.
An extensive overview over MLC is provided by \citet{MultiLabel-Overview}.\tod{include other ref}

The simplest solution to MLC is to learn a binary classifier $h_j$ for each of the labels $\lambda_j$ using the corresponding column in $Y$ as target signal. The approach is referred to as binary relevance decomposition (BR) and disregards dependencies between the labels. 
For instance, BR might predict contradicting label combinations (for a specific dataset) since the labels are predicted  independently from each other. 
\raus{Label powerset (LP) transformation instead is tailored at predicting correct label combinations since it assumes each observed $\vy$ in $Y$ as separate classes in a multi-class classification problem. 
However, LP is not able to predict new label combinations which might for example occur when some labels are independent. 
Moreover, the computational costs might increase exponentially with the number of labels.}

\subsection{(Dynamic) Classifier Chains}
The approach of classifier chains \cite{cc} overcomes the disadvantages of BR as it neither assumes full label independence nor full dependence. 
As in BR, a set of $N$ binary classifiers is trained, but in order to being able to consider dependencies, the classifiers are connected in a chain according to the Bayesian chain rule and pass their predictions along a chain. Each classifier then takes the predictions of all previous ones as additional features and builds a new model.


More specifically each $h_j$ is trained on the augmented training data $X\times Y_{\bullet,1}\times\ldots\times Y_{\bullet,j-1}$ to predict the $j$-th column  $Y_{\bullet,j}$ of $Y$ based on previous predictions $\hat{y_1},\ldots,\hat{y}_{j-1}$ as follows
\begin{align}
\hat{y}_j= h_j(\vx,\hat{y_1},\ldots,\hat{y}_{j-1})
\end{align}
with $\hat{y}_1=h_1(\vx)$ and assuming for convenience an ascending order on the labels.\tod{include phases and var $t$?}

As further research revealed, CCs are able to capture global dependencies as well as  dependencies appearing only locally in the instance space \citep{dembczynski12PCCdependence}. 
However, the ability of the CC approach to capture dependencies is determined by the chosen ordering of the labels. 
A common approach is to set the order of the labels randomly. 
Early experimental results already revealed that the ordering has an obvious effect on the predictive performance \citep{cc, cc_label_ordering2}.
A solution is to use ensembles of CCs with different orderings \citep{cc,secc}, but creating and maintaining an ensemble of CCs is not always feasible~\citep{cc_label_ordering} and comes with further issues on combining the predictions.
\umschreiben{
An alternative way to handle the label ordering problem is to determine a good chain sequence in advance. 
For this purpose methods such as genetic algorithms~\citep{cc_label_ordering}, Bayesian networks~\citep{cc_bayesian} or double Monte Carlo optimization technique
\citep{read2014efficient}
have been used.
}
\tod{use more often "static" ordering}

\com{entweder so, oder noch eine subsection mit dynamic classifier chains, und dann möglicherweise DCC-RDT besser erklären}
Apart from the computational disadvantages of exploring different label sequences, which often leads to just choosing a random ordering in practice, another issue is the underlying assumption that there is one unique, globally optimal ordering which fits equally to all instances. 
Instead, dynamic approaches choose the label ordering depending on the test instance at hand. 
For instance, \citet{cc_label_ordering2} determine the classification order on the fly by looking at the nearest neighbors of the test instance and using the label ordering which works well on the neighbors. However, the approach is computationally expensive since new CC models have to be build during prediction.
\citet{jn:NIPS-17-MLC-RNN} use recurrent neural networks to predict the positive labels as a sequence.
\citet{jn:ICML-19} further use reinforcement learning to determine a different, best fitting sequence over the positive labels of each training instance.
Predicting only the positive labels has the advantage of considerably lower computational costs during prediction, since the number of relevant labels is usually low in comparison to the total number of available labels. 
\drin{The advantage comes at the expense of  ignoring dependencies to negative labels. Predicting the absence of a label is often much easier than finding positive ones and the knowledge about the absence of a label might be very useful to predict a positive label.\tod{include example? the image of moritz is nice, perhaps we have a similar example?}}
Despite the induction of the positive labels does not depend on the number of labels, the approaches of \citet{jn:NIPS-17-MLC-RNN,jn:ICML-19} can still be computationally very demanding due to the complex neural architectures needed, especially regarding the usage of reinforcement learning which actually has again to explore many possible label sequences during training. 

\citet{mk:DS-18-DCC-RDT} propose to integrate dynamic classifier chains in random decision trees (RDT) \citep{rdt_ml}.
In contrast to the common induction of decision trees or to random forests, RDTs are constructed completely at random without following any predictive quality criterion. \tod{talk more about the inner nodes}
\citet{mk:DS-18-DCC-RDT} place tests on the labels at the inner nodes, which they can turn on and off without altering the original target of the RDT since it 
is only specified during prediction by the way of combing the statistics in the leaves.
Hence, it is possible to simulate any binary base classifier of a CC in any possible chain sequence.
In an iterative process, the same RDTs is queried subsequently to determine the next most certain (positive or negative) label.
In the respective next iteration the predicted label is added to the input features like in CC and the respective label tests are turned on. 

The results of the experimental evaluation of DCC-RDT show that the dynamic classification achieves a major improvement over static label orderings. 
However, the lack of any optimization may lead to an insuperable gap to state-of-the-art methods. 
In fact, the results also show that RDT are inherently not suitable for sparse data like text.

\subsection{Extreme Gradient Boosted Trees}
Extreme Gradient Boosted Trees \emph{(XGBoost)} 
\citep{chen2016xgboost}
is a versatile implementation of gradient boosted trees. 
One of the reasons for its success is the very good scalability due to the specific usage of advanced techniques for dealing with large scale data.
XGBoost was originally designed for dealing with regression problems, but different objectives can be defined by correspondingly adapting the objective function and the interpretation of the numeric estimates.
Each model consists of a predefined number of decision trees. These trees are built using gradient boosting, i.e., the model is step-wise adding trees which further minimize the training loss. 
They are constructed recursively, starting at the root node, by adding feature tests on the inner nodes.
At each inner node, all possible feature tests are evaluated according to the gain obtained by applying the split on the data. 
The test candidate returning the highest gain score is then taken and both children are further split up until the maximum depth is reached or the gains stay below a certain threshold. A prediction can be calculated by passing an instance through all trees and summing up their respective leaf scores.

\paragraph{Boosted Optimization}
We refer to 
\citep{chen2016xgboost}
for a more detailed description of XGBoost.
An XGBoost model consists of a sequence of $T$ decision trees $f_1, \ldots, f_T$.
Each tree returns a numeric estimate $f_c(\vx)$ for a given instance $\vx$. 
Predictions are generated by passing an instance through all trees and summing up their leaf scores. The model is trained in an additive manner and each boosting round adds a new tree that improves the model most. For the $t$-th tree the loss to minimize becomes
\begin{align}    
	L^{(t)} = \sum_{i=1}^M l\left(y_i,\ (\hat{y}_i^{(t-1)}+f_t(\vx_i))\right)+\Omega(f_t)
\end{align}
where $l(y,\hat{y})$ is the loss function for each individual prediction, $\hat{y}_i^{(t-1)}=\sum_{k=1}^{t-1} f_k(\vx_i)$ the prediction of the tree ensemble so far and  $\Omega$ is an additional term to regularize the tree.
Combined with a convex differential loss function the objective can be simplified by taking the second-order approximation which gives us the final objective to optimize:
\begin{align}		
	{obj}^{(t)} &= \sum_{v=1}^T[G_v w_v + \frac{1}{2}(H_v +\epsilon)w_v^2] + \gamma T
    \quad \text{with} &G_v = \sum_{i \in I_v} g_i \text{, } H_v = \sum_{i \in I_v} h_i 
\end{align} 
$I_v = \{i|q(x_i)=v\}$ is the set of indices for all data points in leaf $v$. $G_v$ defines the sum of the gradients for all instances $I_v$ in leaf $v$, $H_v$ the corresponding sum of the Hessians and $w_v$ the vector of leaf scores. With the optimal weights $w_v^*$ for leaf $v$ the objective becomes
\begin{align}
		{obj}^* &= -\frac{1}{2} \sum_{v=1}^T \frac{G_v^2}{H_v + \epsilon} + \gamma T \hspace{1cm}\text{with}\hspace{0.7cm} w_v^* = -\frac{G_v}{H_v + \epsilon}
		\label{eq:obj}
\end{align}
These weights finally lead to the gain function used to evaluate different splits. The indices $L$ and $R$ for $G$ and $H$ refer to the proposed right and left child candidates: 
\begin{align}
	L_{split} &= \frac{1}{2}\left[\frac{G_L^2}{H_L + \epsilon} +  \frac{G_R^2}{H_R + \epsilon} - \frac{(G_L + G_R)^2}{H_L + H_R + \epsilon}\right] -\gamma
\label{eq:gain}
\end{align}

There are only few special adaptations of the gradient boosting approach to MLC in the literature and they mainly deal with computational costs. 
Both \citep{pmlr-v70-si17a} and \citep{zhang2019GBDT-MO} propose 
to exploit the sparse label structure which they try to transfer to the gradient and Hessian matrix by using $L$0 regularization.
These approaches are limited to decomposable evaluation measures, which roughly speaking means that, opposed to the classifier chains approaches, they are tailored towards predicting the labels separately rather than jointly.  
Moreover, different technical improvements regarding parallelization and approximate split finding are proposed which could also be applied to the proposed technique in the following.  

\section{Learning a Dynamic Chain of Boosted Tree Classifiers\com{(3 pages)}}
	Instead of learning a static CC that predicts labels in a predefined rigid order, we introduce a dynamic classifier chain (DCC) where each chain-classifier predicts only a single label which is not predetermined and can be different for each sample. To prevent learning bad label dependencies, the base-classifiers are built to maximize the probability of only a single label. On the one hand this allows to exploit more complex label dependencies, and on the other side to massively reduce the length of the chain, while still being able to predict all labels. Given a dataset with 100 labels and a cardinality of five, a DCC of length five has the ability to predict all labels, whereas a CC would have to train 100 models. 
	Because of XGBoosts highly optimized boosting-tree architecture, we decided to use it as base-classifiers in our chain. Therefore we have to modify it and make it capable of building multilabel-trees which can deal with an arbitrary number of labels.

	\subsection{Multi-label XGBoost}
	\label{sec:ML-XGB}
Since XGBoost only supports binary classification with its trees in the original implementation, the underlying tree structure had to be adapted in order to support multi-label targets. 

The first modification is to calculate leaf weights and gradients over all class labels instead of only a single one. 
More specifically, $G_{j,v} = \sum_{i \in I_v} g_{j,i}$ and $H_{j,v} = \sum_{i \in I_v} h_{j,i}$ extend to the labels $1\leq j \leq N$.
In consequence, the objective \eqref{eq:obj} and gain functions \eqref{eq:gain} have to be adapted to consider gradient and hessian values from all classes. 
A common approach in multi-variate regression and multi-target classification is to compute the average loss of the model over all targets \citep{waegeman19multitarget}. 
Adapted to our XGBoost trees, this corresponds to the sum of $\frac{G_j^2}{H_j+\epsilon}$ over all labels (cf. Table~\ref{tab:splitmethods}). We refer to it as the \textbf{\ap{sumGain}} split method.
We use cross entropy as our loss, as it has demonstrated to be appropriate practically and also theoretically for binary and especially  multi-label classification tasks \citep{nam14revisiting,dembczynski12PCCdependence}.
Hence, the loss is computed as (shown here only for a single label)
\begin{align}
    l(y,\hat{y}')=-y\log(\hat{y}')+(1-y)\log(1-\hat{y}')
\end{align}
In order to get $\hat{y}'$ as a probability between zero and one, a sigmoid transformations has to be applied to the summed up raw leaf predictions $\hat{y}=\sum_{t=1}^{T} f_t(\vx)$, returned from all boosting trees, where $\hat{y}' = sigmoid(\hat{y}) = \frac{1}{1+e^{-\hat{y}}}$. This is also beneficial for calculating $g$ and $h$, since the gradients of the loss function simply become
\begin{align}\label{eq:gh}
    g = \hat{y} - y \hspace{0.7cm}\text{and}\hspace{0.7cm} h = \hat{y} \cdot (1-\hat{y})
\end{align}

One might not expect a very different predictions from the combined formulation than from minimizing the loss for each label separately by separate models (as by BR)\com{Fehlt da ein Wort? elm:jetzt?}. 
However, as \citet{waegeman19multitarget} note fitting one model to optimize the average label loss has a regularization effect that stabilizes the predictions, especially for infrequent labels. 
In addition, only one model has to be inferred in comparison to $N$ which has a major implication on the computational costs. 
This is especially an advantage in the case of a large number of labels and our proposed dynamic approach can directly benefit from it.\com{gefaellt mir nicht ganz letzter satz}

	\begin{figure}[tb]
		\centering
		\resizebox{\textwidth}{!}{\includegraphics{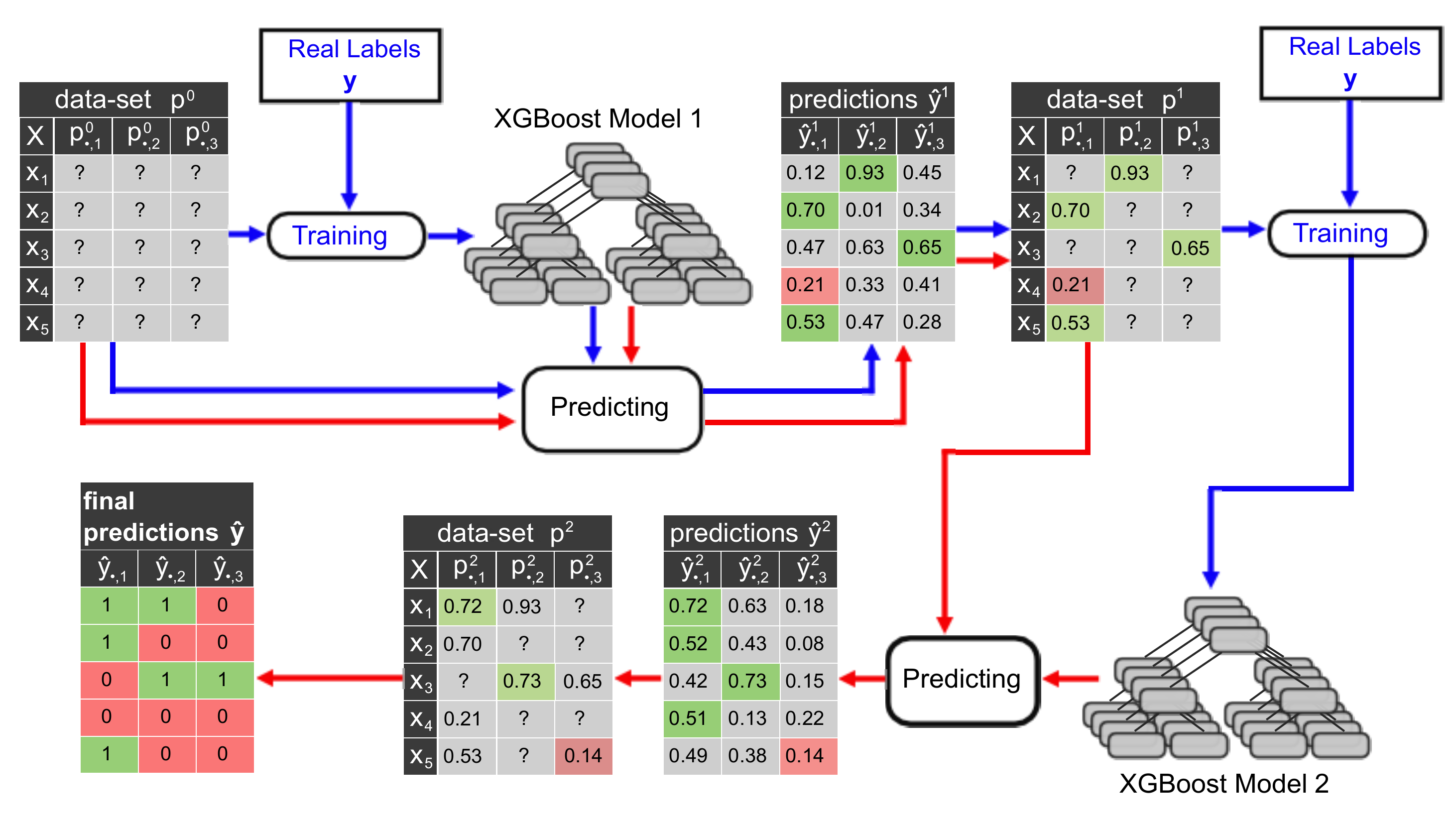}}
		\caption{Dynamic Chain: training pipeline (blue arrows) \& prediction pipeline (red arrows).}
		\label{fig:chain-train}
	\end{figure}

\subsection{Extreme Dynamic Classifier Chains}\tod{Überleitung von MultiTarget Trees zu der Kette und besonders den Split Methoden} 
\label{sec:XDCC}
After introducing the ML-XGBoost models, which can deal with multiple labels, the next step is to modify the tree construction to align it with our goal of predicting a single label per instance.

\begin{table}[tb]
\caption{Proposed split gain calculations with a simplified example calculation for the predicted scores $\hat{y}=(0.8,0.2,0.9,0.1)$ of the previous trees  and given true labels $y=(1,1,0,0)$. For convenience, we  assume $H_j+\epsilon=1$.}
\begin{tabular*}{\textwidth}{c @{\extracolsep{\fill}} ccccc}
\hline
Gain & Formula & Example & Gain & Formula & Ex. \\
\hline
\ap{sumGain} & $\displaystyle\sum_{j=1}^N \left(\frac{G_j^2}{H_j+\epsilon}\right)$ 
& $0.2^2+0.8^2+0.9^2+0.1^2$ &
\ap{maxGain} & $\displaystyle\max_{1 \leqslant l \leqslant N}\left(\frac{G_j^2}{H_j+\epsilon}\right)$ 
& $0.9^2$ \\
\ap{sumWeight} & $\displaystyle\sum_{l=1}^N \left(\frac{-G_j}{H_j+\epsilon}\right)$ 
& $0.2+0.8-0.9-0.1$ &
\ap{maxWeight} & $\displaystyle\max_{1 \leqslant l \leqslant N}\left(\frac{-G_j}{H_j+\epsilon}\right)$ 
& $0.8$ \\
\ap{sumAbsG} & $\displaystyle\sum_{l=1}^N \left(\left|\frac{-G_j}{H_j+\epsilon}\right|\right)$
& $0.2+0.8+0.9+0.1$ &
\ap{maxAbsG} & $\displaystyle\max_{1 \leqslant l \leqslant N}\left(\left|\frac{-G_j}{H_j+\epsilon}\right|\right)$
& $0.9$ \\
\end{tabular*}
\label{tab:splitmethods}
\end{table}

\tod{motivate further that we have different methods. some of the attain to our algorithmic design (the max variants (find only one), and the weight variants (only positive)), but why should we then apply all of them if e.g. we want to use a chain. one argument could be that it might still be data dependent (there might be stronger dependencies from negative to positive labels. due to cumulated predictions it might be good to have good overall predictions from the beginning), so we prefer to use it as a hyperparameter}\res{Hab zu den Varianten je kurz einen Satz geschrieben. Darunter kommt es eigentlich aber nochmal.}
Table~\ref{tab:splitmethods} 
shows the proposed split functions and an example for each one to demonstrate the gain calculations. We assume to have a single instance with four different target labels $y$ and their corresponding predictions $\hat{\vy}$. $g$ and $h$ are calculated according to Eq.~\eqref{eq:gh} and we get $G=(-0.2,-0.8,0.9,0.1)$. 
\drin{We have focused on different characteristics for each function. \com{Brauchen wir den Teil von hier bis ...}The \textit{max} versions focus on optimizing a tree for predicting only a single label, whereas \textit{sum} functions aim for finding a harmonic split that generates predictions with high probabilities over all labels. The \textit{weight} variants focus on directly optimizing the tree outputs and hence prefer positive labels, while the \textit{gain} splits stay close to XGBoosts original gain calculation and try to optimize positive and negative labels to the same extend. \com{... hier? oder auch das danach?}}
Hereinafter we give a more detailed description and motivation for each gain function:
\begin{description}
\item[Maximum default gain over all labels]  
XDCC predicts labels one by one. It hence does not need to find a split which increases the expected loss over all labels (such as \ap{sumGain}), but only one.
Hence, \textbf{\ap{maxGain}} is tailored to find the label with maximal gain, which corresponds to the label for which the previous trees produced the largest error. 
In the example in Table~\ref{tab:splitmethods}, this corresponds to $\lambda_3$ for which a change of $0.9^2$ w.r.t. cross entropy was computed if the prediction is changed to the correct one.
\item[Sum and maximum gradients over all labels] 
In contrast to \ap{maxGain}, \textbf{\ap{sumGrad}} aims at good predictions for positive labels only and hence corresponds to the idea of predicting the positive labels first. 
Positive labels obtain positive scores, whereas negative labels obtain negative scores. The variant \textbf{\ap{maxGrad}} chooses the  positive label for which the greatest improvement is possible and only goes for the best performing negative label if there are no true positive labels in the instance set.
In the example, $\lambda_2$ is chosen since the improvement is greater than for $\lambda_1$, and definitely greater as for the negative labels.
\item[Sum and maximum absolute gradients over all labels] 
In contrast to \ap{sumGain} and \ap{maxGain}, the measures \ap{sumGrad} and \ap{maxGrad} not only favour positive labels but also take the gradients linearly instead of quadratically into account. 
This might, for instance,  reduce the sensitivity to outliers. 
Hence, we also include two variants \textbf{\ap{sumAbsG}} and \textbf{\ap{maxAbsG}} which encourage to predict the labels where the model would improve the most, regardless whether it is positive and negative, but which similarly to \ap{sumGrad} and \ap{avgGrad} use a linear scale on the gradients.
\end{description}
Even though DCC's original design is to predict a single positive label per round, 
good overall predictions might be required from the beginning for instance in the case of shorter chains. 
Therefore, we use the split-method as an additional hyperparameter to choose it individually for different XDCC variants and datasets.

	\subsubsection{Training Process}\tod{was noch irgendwo fehlt ist, dass mehrere baeume pro runde moeglich sind. kann man vlt erschlagen, indem man irgendwo $\hat{y}$ konkret definiert, also $\hat{y}^r=\sum_{t=1}^{T} f^r_t(\x)$ oder so. mittlerweile hab ich es an verschiedenen stellen eingefügt}
	A schematic view for training the dynamic chain with a length of two is shown in Figure~\ref{fig:chain-train} following the blue lines.
	In a first preprocessing step, the training datasets have to be adapted. For each label $\lambda_j$ a new \emph{label-feature} $p_j^0$, initialized as unknown (?), is added to the original features resulting in the augmented space $\vx,\vp \in X \times [0,1]^N$. While proceeding through the chain, these "?" values are replaced with predicted label probabilities 
	out of	$\hat{\vy}^r=\sum_{t=1}^{T} f^r_t((\vx,\vp^{r-1}))$
	in round $r$. \res{Wollte hier noch das mit den Bäumen einfügen. Notwendg? They are generated by evaluating all boosted trees of the model in round $r$ by calculating $\hat{\vy}^r=\sum_{t=1}^{T} f^r_t((X,\vp^{r-1}))$}. As soon as these feature columns begin to be filled with values, following classifiers may detect dependencies and base their predictions on them. 
	Each round $r$, for $1\le r \le N$, starts with training a new ML-XGBoost model by passing the train set combined with the additional label-features $\vp^{r-1}$ and the target label matrix $\vy$ to it. Afterwards, the model is used to generate predictions $\hat{\vy}^r$ on the same data used to train it, shown in the \textit{predictions} tables. In the last step these predictions are then propagated to the next chain-classifier, by replacing the corresponding label features with the predicted probabilities $\vp^{r}$. Three different cases can occur during this process:
	\begin{itemize}
	    \item At least one label, that was not propagated previously, has a probability $\ge 0.5$: The label with the highest probability is propagated.
	    \item All labels, that were not propagated previously, have probabilities $< 0.5$: The label with the lowest probability is propagated.
	    \item Otherwise, no additional label is propagated.\com{Den letzten Punkt verstehe ich nicht.}\res{So in Ordnung? Will sagen: $p_{i,j}^r = p_{i,j}^{r-1}$}
	\end{itemize}
	They can be formalized where $p_{i,j}^r$ denotes the added label feature and $p_{i,j}^r$ the corresponding predictions for label $\lambda_j$ of an instance $x_i$ in training round $r$.
    \begin{align}
	  p_{i,j}^r =
   \begin{cases}
     \hat{y}_{i,j}^r    & \text{if } p_{i,j}^{r-1} = ? \text{ and }  \max_m \hat{y}_{i,m}^r \geq 0.5 \text{ and }   \hat{y}_{i,j}^r = \max_m \hat{y}_{i,m}^r \\
     \hat{y}_{i,j}^r    & \text{if } p_{i,j}^{r-1} = ? \text{ and }   \max_m \hat{y}_{i,m}^r < 0.5 \text{ and } \hat{y}_{i,j}^r = \min_m \hat{y}_{i,m}^r \\
     p_{i,j}^{r-1}      & \text{otherwise } 
   \end{cases}
	\end{align}
    In all cases where labels are propagated, later classifiers are not allowed to change these labels from positive to negative or the other way around, based on the assumption that later classifiers tend to have a higher error rate, since their decisions are based on previous predictions \citep{senge2014problem}.

	\subsubsection{Prediction Process}
	The prediction process is similar to the training process. Instead of training a model in each step, we reuse the models from the training phase to generate predictions on the test set. After all predictions are propagated, the propagated labels are mapped to label predictions, where probabilites $p_{i,j} < 0.5$ or equal to $?$ are interpreted as negative labels and probabilities $p_{i,j} \ge 0.5$ as positive labels. The process is depicted in Figure~\ref{fig:chain-train} following the red lines.

	\subsection{Refinements to the chain}
	In this section we shortly describe two problems we faced during development of the DCC approach and propose methods to tackle them.

	\subsubsection{Separate and Conquer}
	Consecutive models in the chain tend to select the same splits and therefore predict the same labels, especially ones which are easy to learn, i.e. if they clone existing features. We solve this problem by introducing an approach similar to separate-and-conquer from rule learning \citep{jf:AI-Review}. The \textit{separating} step turns all gradient and hessian values of previous predicted labels for an instance to zero. Thereby they are no longer considered during split score calculation in the \textit{conquering} step and other splits become more likely since scores for already used splits are lower.

	\subsubsection{Cumulated Predictions}
	A second observation was that final predictions, after traversing the chain, contain too little positive labels. Analyzing the chain models showed that especially early models predict multiple positive labels, but are only allowed to propagate the one with the highest probability. Therefore we introduce \textit{cumulated predictions} to preserve these otherwise forgotten positive predictions.
	The idea is to save all predictions of each chain classifier and merge them afterwards with the chain predictions of the unmodified DCC using the following heuristic. 
	The final cumulated prediction $c_{i,j}$ for label $\lambda_j$ and instance $\vx_i$ is computed as
    \begin{align}
	  c_{i,j} = 
   \begin{cases}
    p_{i,j}^N                                   &\text{if $p_{i,j}^N \ne$ ?} \\
    \max(\hat{y}_{i,j}^1, ..., \hat{y}_{i,j}^N)      &\text{otherwise} 
   \end{cases}
	\end{align}

\section{Experiments}
The experiments were evaluated for the following models, where BR, CC and RDT serve as baselines:
\begin{itemize}
\item \textbf{BR}: Binary Relevance with default XGBoost models for binary-classification. 
\item \textbf{CC}: Classifier Chains with a random order and default XGBoost as base-models.
\item \textbf{RDT-DCC}: Dynamic Classifier Chains using Random Decision Trees \citep{mk:DS-18-DCC-RDT}.
\item \textbf{ML-XGB}: A single multi-label XGBoost model introduced in Section~\ref{sec:ML-XGB}. 
\item \textbf{XDCC}: Our Dynamic Classifier Chain with ML-XGB models as base classifiers.
\item \textbf{XDCC cumulated}: The cumulated version of DCC introduced in Section~\ref{sec:XDCC}.
\end{itemize}

\begin{table}[t!]
\caption{Datasets, \# of instances, labels, cardinality, \# of distinct label combinations.}
\label{tab:datasets}
\centering
\resizebox{\textwidth}{!}{
\begin{tabular}{lcccc|lcccc}
\toprule
dataset	&	instances	&	 labels	&	cardinality	&	distinct 	&	dataset	&	instances	&	 labels	&	cardinality	&	distinct \\
\midrule
\ds{emotions}	&	593	&	6	&	1.869	&	27	&	\ds{genbase}	&	662	&	27	&	1.252	&	32 \\
\ds{scene}	&	2407	&	6	&	1.074	&	15	&	\ds{medical}	&	978	&	45	&	1.245	&	94 \\
\ds{flags}	&	194	&	7	&	3.392	&	54	&	\ds{enron}	&	1702	&	53	&	3.378	&	753 \\ 
\ds{yeast}	&	2417	&	14	&	4.237	&	198	&	\ds{bibtex}	&	7395	&	159	&	2.402	&	2856\\
\ds{birds} 	&	645	&	19	&	1.014	&	133	&	CAL500 	&	502	&	174	&	26.044	&	502 \\
\ds{tmc2007}	&	28596	&	22	&	2.158	&	1341	&		&		&		&		&	\\
\bottomrule
\end{tabular}
}
\end{table}

The evaluated datasets in Table~\ref{tab:datasets} cover a wide variety of application areas for multi-label classification.
The number of labels vary between a few (6) and hundreds (374). The number of labels per instance (cardinality) is usually low in comparison ($<$10), but can raise up to 26 (\ds{CAL500}).
All datasets came with predefined train-tests splits which were used for the final evaluation. 
Parameters were tuned in terms of obtaining best F1 on randomly chosen validation set containing 20\% of the train set.%
\footnote{
The following parameters were tuned by grid-search:
number of trees \{100, 300, 500\}, 
max tree depth \{5, 10, 20, 30, 50\},  \com{der Parameter ist doppelt, weil RDT hier 30 prüft und die anderen Modelle 100; zusammenfassen? elm: ne egal}
percentage labels \{0.1, 0.2, 0.3\}  for RDT,
max tree depth \{5, 10, 20, 50, 100\}, 
number of boosting rounds \{10, 20, 50, 100\},
learning rate \{0.1, 0.2, 0.3\} for XDCC, ML-XGB, BR, CC, 
split methods in Table~\ref{tab:splitmethods} for XDCC, ML-XGB.
}

\subsection{Evaluation Measures}
\label{sec:measures}
From the large variety of evaluation measures that exist for MLC the most interesting ones for analyzing our proposed methods are \emph{Hamming accuracy} (HA) and \emph{subset accuracy} (SA).
\drin{Hamming accuracy denotes the accuracy of predicting individual labels averaged over all labels, whereas 
subset accuracy measures the ability of a classifiers of predicting exactly the true label combination for an instance. }
\drin{In the case of predicting a large amount of labels, subset accuracy is often of limited use since it often evaluates to zero. 
Hence, we additionally consider \emph{example-based F1} as measure especially for the parameter tuning.
It can be considered as a compromise between Hamming and subset accuracy and was also used by \citet{jn:ICML-19,jn:NIPS-17-MLC-RNN} as surrogate loss for subset accuracy.}
\raus{We also include \emph{example-based F1} as a compromise between both measures \citep{jn:NIPS-17-MLC-RNN,jn:ICML-19}, and hence also as objective for the parameter tuning.}
More formally, the comparison between true $\vy$ and predicted $\hat{\vy}$ for a test instance $\vx$ is evaluated to
(with $\mathbb{I}$ as indicator function)
\begin{equation*}
\text{SA}= \mathbb{I}\left[\vy = \hat{\vy}\right] \quad 
\text{HA}= \frac1N \sum_{j=1}^N \mathbb{I}\left[y_j = \hat{y}_j \right] \quad
\text{F1}=\frac{2 \sum_{j=1}^N y_j \hat{y}_j}{ \sum_{j=1}^N y_j + \sum_{j=1}^N \hat{y}_j} 
\end{equation*}

\raus{HA is decomposable with respect to  the labels, whereas SA is not.} 
As \citet{dembczynski12PCCdependence} indicate, HA and SA  are orthogonal to each other. 
From a probabilistic perspective, to predict the true label combination requires to find the mode of the joint label distribution, whereas it is sufficient to find the modes of the marginal label distributions for Hamming accuracy. 
If there are dependencies between labels, both modes do not have to coincide. 
\drin{In consequence, an approach such as BR is sufficient if one is interested in good HA (or there are no dependencies).  
CC\drin{, especially if using the same base learner and configuration as its BR counterpart,} cannot be expected to improve over BR regarding Hamming. 
On the other hand, the reverse behaviour can be expected for SA. }
Hence, the trade-off between both measures and the relation to BR can serve us to assess the ability of considering label dependencies.

\begin{figure}[tb!]
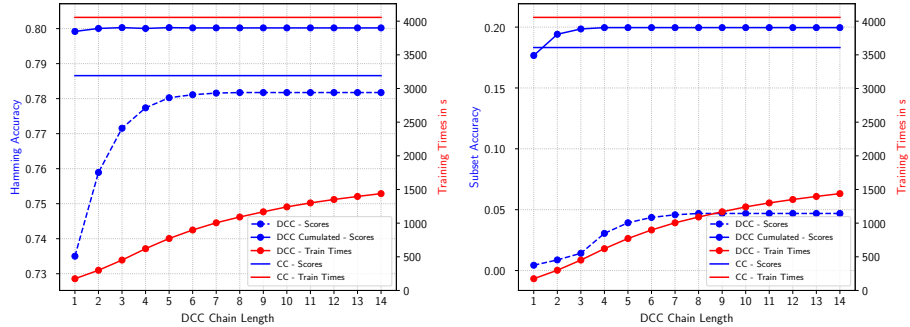

\centering
\begin{subfigure}{.5\textwidth}
    \scalebox{.38}{\input{figures/yeast_HammingAcc_opt-SXGB.pgf}}
\end{subfigure}%
\begin{subfigure}{.5\textwidth}
    \scalebox{.38}{\input{figures/yeast_SubsetAcc_opt-SXGB.pgf}}
\end{subfigure}%
\caption{Comparison with respect to length of the chain on \ds{yeast}  wrt. Hamming  and subset accuracy. 
}
\label{fig:performance-yeast}
\end{figure}
\raus{
\begin{figure}[tb!]
\centering
\begin{subfigure}{.5\textwidth}
    \scalebox{.38}{\input{figures/tmc2007_F1_opt-SXGB.pgf}}
\end{subfigure}%
\begin{subfigure}{.5\textwidth}
    \scalebox{.38}{\input{figures/tmc2007_SubsetAcc_opt-SXGB.pgf}}
\end{subfigure}%
\caption{Comparison with respect to length of the chain on \ds{tmc2007} wrt. F1 and subset accuracy. 
}
\label{fig:performance-tmc}
\end{figure}
}

\begin{figure}[tb!]
    \centering
    \resizebox{\textwidth}{!}{
    \includegraphics{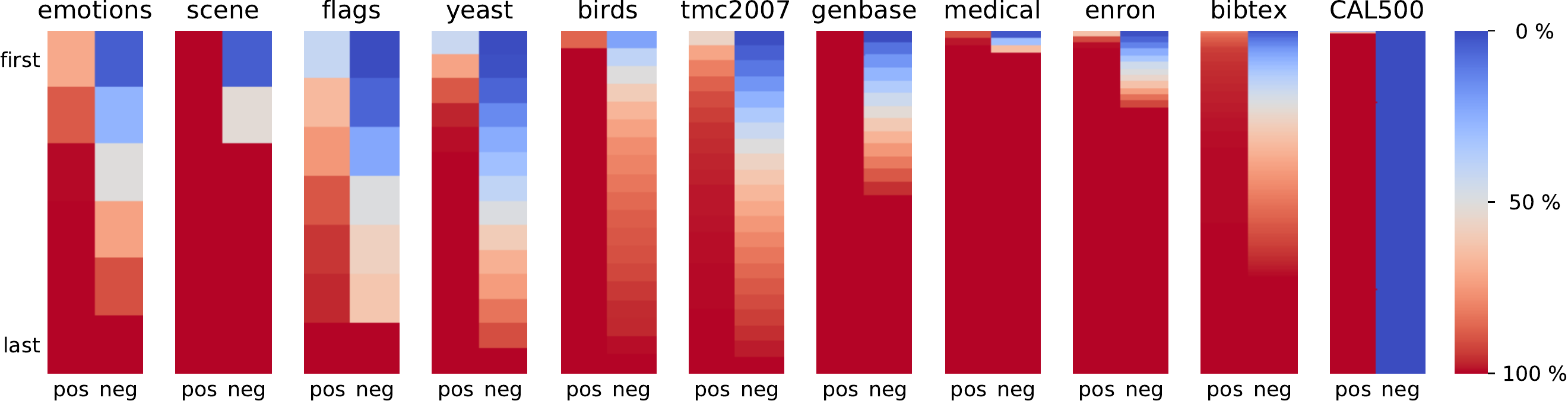}
    }
    \caption{Heat maps of the development of the predictions of positive  and negative labels, left and right side of the bar, respectively, from the first to last round given as fraction of the total number of positive and negative predictions (last row, respectively). 
    }
    \label{fig:prediction_development}
\end{figure}

\subsection{Results}

As described in Section~\ref{sec:XDCC}, XDCC can provide a meaningful prediction after each round, which is a major advantage over CC in terms of computational costs. 
Moreover, by subsequently refining its predictions based on previous predictions, we expected to advance especially in terms of SA. 
Figure~\ref{fig:performance-yeast} shows measures HA, SA and training times for different lengths of the chain on \ds{yeast}. 
CC and XDCC were trained with optimal parameters for CC for a fair comparison of the computational times. Note that length 1 corresponds to ML-XGB when the same parameter were used. 
The first observation is that, as expected, HA and SA increase with increasing length for the standard XDCC variant until a little bit further than the average cardinality of 4 of the dataset. 
If we add the cumulated predictions, the performances converge much faster. Yet, there is a clear improvement visible for SA, which indicates that XDCC$_{cum}$ is able to directly benefit from the previous predictions in order to match the correct label combinations. 
The cumulated predictions are also decisive to surpass CC. 
Interestingly, the training costs of CC are also never reached although the same XGBoost parameters were used.
\raus{
This is different on \ds{tmc2007}, as depicted in Fig.~\ref{fig:performance-tmc}. 
On this dataset, training four rounds of XDCC takes as long as building the complete CC chain. 
We can also observe that although the improvement with respect to subset accuracy along the chain is not sufficient to reach the performance of CC in this case, adding rounds leads to an improvement in terms of example-based F1, which can be considered as trade-off between the two extreme measures subset and Hamming accuracy. 
}

\begin{table}[tb]
  \caption{Predictive performance and training times comparison. Shown are the average ranks over the 10 datasets and the ranks over these in brackets.\tod{If time, make ranks of ranks smaller and put more space between columns.} }
  \label{tab:overallcomparison}
\centering
    \begin{tabular}{lcccc}
\toprule
Method & \hspace{2em}HA\hspace{2em} & \hspace{2em}SA\hspace{2em} & \hspace{2em}F1\hspace{2em} & \hspace{0em}Train time\hspace{0em}  \\
\midrule
BR & 2.20 (1) & 3.45 (4) & 3.00 (2) & 3.00 (3) \\
CC & 3.05 (3) & 2.60 (1) & 3.30 (3) & 3.30 (4) \\
RDT-DCC & 5.10 (6) & 4.05 (5) & 3.60 (5) & --  \\
\midrule
ML-XDB  & 2.90 (2) & 3.20 (3) & 3.35 (4) & 1.10 (1)  \\
XDCC$_{cum}$ & 3.15 (4) & 3.05 (2) & 2.45 (1) & 2.60 (2) \\
XDCC$_{std}$ & 4.60 (5) & 4.65 (6) & 5.30 (6) & " \\
\bottomrule
    \end{tabular}
\end{table}
\begin{figure}[b!]
    \centering
    \resizebox{0.7\textwidth}{!}{\includegraphics[trim={0cm 0.1cm 0cm 1.15cm},clip]{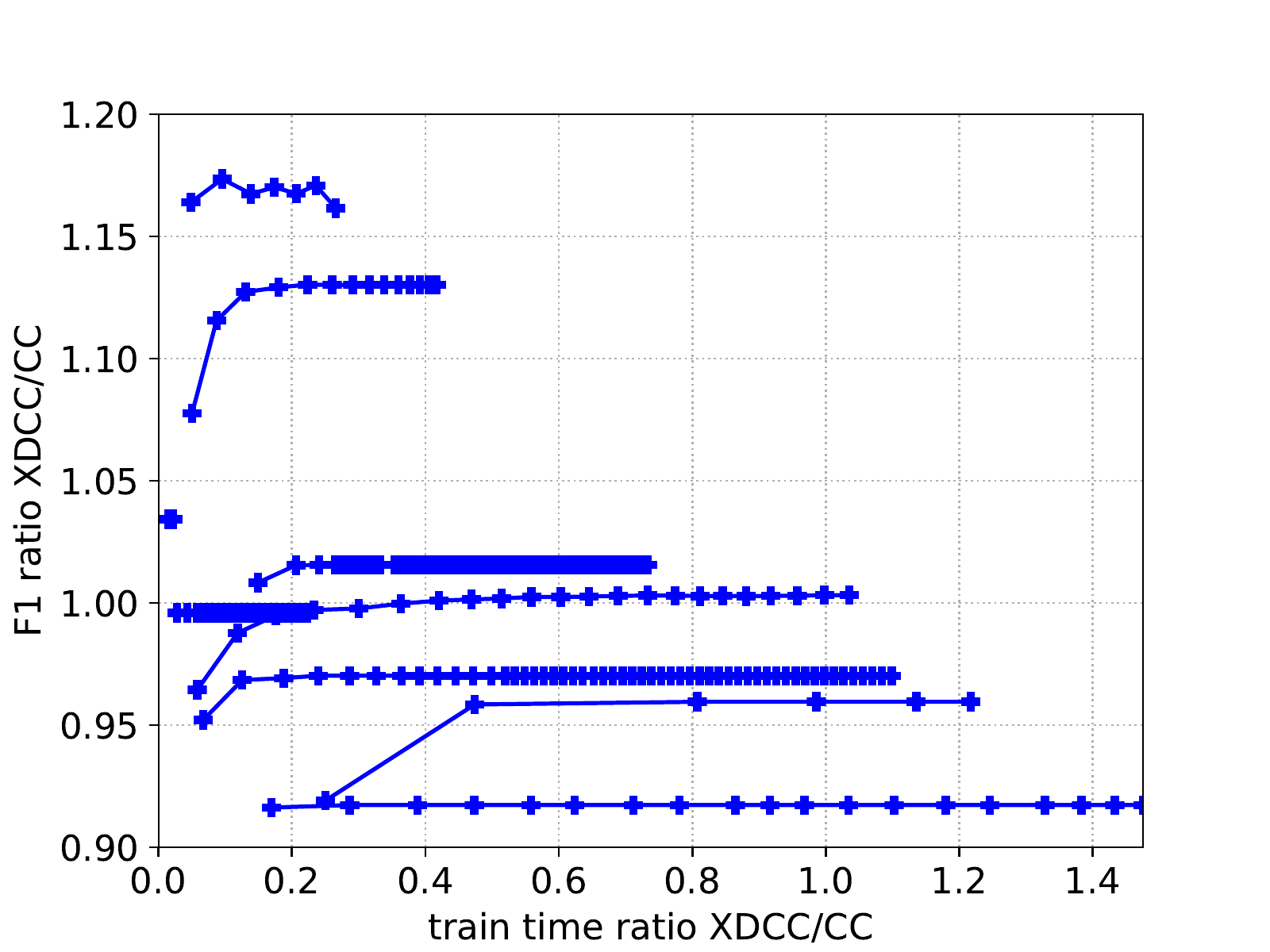}}
  \caption{Train time ratios between XDCC$_{cum}$ and CC in relation to their ratio with respect to F1 for nine datasets. For instance, all points below $x=1$ and $y=1$ indicate XDCC$_{cum}$ models  which consume less training time but perform worse than CC w.r.t. F1. }
  \label{fig:ratios}
\end{figure}

The point where train times of CC are reached by XDCC are further investigated in Figure~\ref{fig:ratios}. It shows the ratio of XDCC$_{cum}$  to CC for the different datasets (connected lines) and chain lengths. \ds{CAL500} around (0.1,1.4) is not shown for convenience.
The diagram shows that XDCC only takes longer than CC on four datasets and only for the last rounds. 
For three of these datasets XDCC does not reach CC's F1. 
As already shown in Figure~\ref{fig:performance-yeast}, XDCC$_{cum}$ only improves in the  first rounds, and sometimes there is even a tendency to decrease.
The progress of predicting the labels is also depicted in Figure~\ref{fig:prediction_development}. 
It visualizes that  positive labels are generally predicted in earlier rounds, as expected from the design of the split functions. 
As shown previously, this behaviour is decisive for the fast convergence and hence the possibility to end the training and prediction processes already in early rounds. 

Table~\ref{tab:overallcomparison} also includes a comparison to the RDT-DCC baseline. 
The first observation is the strong baseline achieved by BR regarding Hamming, as partially expected in Section~\ref{sec:measures}.
In the same way, CC is best in terms of SA. 
However, ML-XGB performs second regarding HA and XDCC$_{cum}$ is second regarding SA, which suggests that the proposed approach is able to trade-off between both extremes. This is also confirmed by the best position in terms of F1. 
RDT is the worst performing approach and is even sometimes surpassed by XDCC$_{std}$ which is only included for 
showing the effect of cumulative predictions.

\section{Conclusions}
We have proposed in this work XDCC, an adaptation of extreme gradient boosted trees which integrate dynamic chain classifier. 
XDCC predicts labels along the chain in a dynamic order which adapts to each test instance individually.
It was shown that the positive labels are predominantly predicted at the beginning of the process, which allows XDCC to achieve its maximum  performance already after a few rounds. 
This allows XDCC to reduce the length of the chain, which together with the multi-target formulation of XDCC leads to substantial improvements in comparison to binary relevance and classifier chains regarding computational costs, often even if the full chain is processed. 
The length of the chain also trades-off between the two orthogonal objectives of BR and CC, leading to in average the best results in terms of example-based F1.
\tod{if space, future work: a lot of parameters, refine details of algorithm, scale to many labels}

We will consider in the future to specifically adapt our approach to the setting of large number of labels, e.g. by integrating some of the sparse techniques proposed in \citep{pmlr-v70-si17a,zhang2019GBDT-MO}. 
Since the number of associated labels per instance is usually not affected by the increasing number of labels, it will be interesting to see how XDCC will behave with respect to computational costs, but also regarding the exploitation of label dependencies since the size of the (dependency) chains should remain of the same size. 
In order to actually benefit computationally from these short chains, we are planning to include a virtual label which indicates the end of the training and prediction process, similar to the idea of the calibrating label in pairwise learning \citep{jf:Neurocomputing}.



%
%
%
\footnotesize
\bibliographystyle{splncsnat}
\bibliography{references2,bib,mlc}

\begin{thebibliography}{20}
\providecommand{\natexlab}[1]{#1}
\providecommand{\url}[1]{\texttt{#1}}
\providecommand{\urlprefix}{}

\bibitem[{Chen and Guestrin(2016)}]{chen2016xgboost}
Chen, T., Guestrin, C.: Xgboost: A scalable tree boosting system.
\newblock In: Proc. of the 22nd SIGKDD Int. Conf. on Knowledge Discovery and
  Data Mining. pp. 785--794. ACM (2016)

\bibitem[{Dembczy\'nski et~al.(2012)Dembczy\'nski, Waegeman, Cheng, and
  H{\"u}llermeier}]{dembczynski12PCCdependence}
Dembczy\'nski, K., Waegeman, W., Cheng, W., H{\"u}llermeier, E.: On label
  dependence and loss minimization in multi-label classification.
\newblock Machine Learning 88(1-2), 5--45 (2012)

\bibitem[{F{\"{u}}rnkranz(1999)}]{jf:AI-Review}
F{\"{u}}rnkranz, J.: Separate-and-conquer rule learning.
\newblock Artificial Intelligence Review 13(1), 3--54 (February 1999)

\bibitem[{Goncalves et~al.(2013)Goncalves, Plastino, and
  Freitas}]{cc_label_ordering}
Goncalves, E.C., Plastino, A., Freitas, A.A.: {A Genetic Algorithm for
  Optimizing the Label Ordering in Multi-label Classifier Chains}.
\newblock In: Proceedings of the IEEE 25th International Conference on Tools
  with Artificial Intelligence. pp. 469--476 (2013)

\bibitem[{Kulessa and Loza~Menc{\'{\i}}a(2018)}]{mk:DS-18-DCC-RDT}
Kulessa, M., Loza~Menc{\'{\i}}a, E.: Dynamic classifier chain with random
  decision trees.
\newblock In: Proceedings of the 21st International Conference of Discovery
  Science (DS-18) (2018)

\bibitem[{Li and Zhou(2013)}]{secc}
Li, N., Zhou, Z.: {Selective Ensemble of Classifier Chains}.
\newblock In: Multiple Classifier Systems: 11th International Workshop on
  Multiple Classifier Systems, pp. 146--156 (2013)

\bibitem[{Loza~Menc{\'{\i}}a et~al.(2010)Loza~Menc{\'{\i}}a, Park, and
  F{\"{u}}rnkranz}]{jf:Neurocomputing}
Loza~Menc{\'{\i}}a, E., Park, S.H., F{\"{u}}rnkranz, J.: Efficient voting
  prediction for pairwise multilabel classification.
\newblock Neurocomputing 73(7-9), 1164 -- 1176 (Mar 2010)

\bibitem[{Nam et~al.(2014)Nam, Kim, Loza~Menc{\'{\i}}a, Gurevych, and
  F{\"{u}}rnkranz}]{nam14revisiting}
Nam, J., Kim, J., Loza~Menc{\'{\i}}a, E., Gurevych, I., F{\"{u}}rnkranz, J.:
  Large-scale multi-label text classification - revisiting neural networks.
\newblock In: Proceedings of the European Conference on Machine Learning
  (ECML-PKDD-14). pp. 437--452 (2014)

\bibitem[{Nam et~al.(2019)Nam, Kim, Loza~Menc{\'{\i}}a, Park, Sarikaya, and
  F{\"{u}}rnkranz}]{jn:ICML-19}
Nam, J., Kim, Y., Loza~Menc{\'{\i}}a, E., Park, S., Sarikaya, R.,
  F{\"{u}}rnkranz, J.: Learning context-dependent label permutations for
  multi-label classification.
\newblock In: Proceedings of the 36th International Conference on Machine
  Learning (ICML-19). pp. 4733--4742 (2019)

\bibitem[{Nam et~al.(2017)Nam, Loza~Menc{\'{\i}}a, Kim, and
  F{\"{u}}rnkranz}]{jn:NIPS-17-MLC-RNN}
Nam, J., Loza~Menc{\'{\i}}a, E., Kim, H.J., F{\"{u}}rnkranz, J.: Maximizing
  subset accuracy with recurrent neural networks in multi-label classification.
\newblock In: Advances in Neural Information Processing Systems 30 (NIPS-17).
  pp. 5419--5429 (2017)

\bibitem[{Read et~al.(2014)Read, Martino, and Luengo}]{read2014efficient}
Read, J., Martino, L., Luengo, D.: Efficient {Monte Carlo} methods for
  multi-dimensional learning with classifier chains.
\newblock Pattern Recognition 47(3), 1535 -- 1546 (2014)

\bibitem[{Read et~al.(2011)Read, Pfahringer, Holmes, and Frank}]{cc}
Read, J., Pfahringer, B., Holmes, G., Frank, E.: Classifier chains for
  multi-label classification.
\newblock Machine Learning 85(3), 333--359 (2011)

\bibitem[{Senge et~al.(2014)Senge, Del~Coz, and
  H{\"u}llermeier}]{senge2014problem}
Senge, R., Del~Coz, J.J., H{\"u}llermeier, E.: On the problem of error
  propagation in classifier chains for multi-label classification.
\newblock In: Data Analysis, Machine Learning and Knowledge Discovery, pp.
  163--170 (2014)

\bibitem[{Si et~al.(2017)Si, Zhang, Keerthi, Mahajan, Dhillon, and
  Hsieh}]{pmlr-v70-si17a}
Si, S., Zhang, H., Keerthi, S.S., Mahajan, D., Dhillon, I.S., Hsieh, C.J.:
  Gradient boosted decision trees for high dimensional sparse output.
\newblock In: Proceedings of the 34th International Conference on Machine
  Learning. Proceedings of Machine Learning Research, vol.~70, pp. 3182--3190
  (06--11 Aug 2017)

\bibitem[{Silva et~al.(2014)Silva, Gon{\c{c}}alves, Plastino, and
  Freitas}]{cc_label_ordering2}
Silva, P.N.d., Gon{\c{c}}alves, E.C., Plastino, A., Freitas, A.A.: {Distinct
  Chains for Different Instances: An Effective Strategy for Multi-label
  Classifier Chains}, pp. 453--468 (2014)

\bibitem[{Sucar et~al.(2014)Sucar, Bielza, Morales, Hernandez-Leal, Zaragoza,
  and Larrañaga}]{cc_bayesian}
Sucar, L.E., Bielza, C., Morales, E.F., Hernandez-Leal, P., Zaragoza, J.H.,
  Larrañaga, P.: {Multi-label Classification with Bayesian Network-based Chain
  Classifiers}.
\newblock Pattern Recognition Letters 41, 14--22 (2014)

\bibitem[{Tsoumakas and Katakis(2007)}]{MultiLabel-Overview}
Tsoumakas, G., Katakis, I.: Multi-label classification: An overview.
\newblock International Journal of Data Warehousing and Mining 3(3), 1--17
  (2007)

\bibitem[{Waegeman et~al.(2019)Waegeman, Dembczy{\'{n}}ski, and
  H{\"u}llermeier}]{waegeman19multitarget}
Waegeman, W., Dembczy{\'{n}}ski, K., H{\"u}llermeier, E.: Multi-target
  prediction: a unifying view on problems and methods.
\newblock Data Mining and Knowledge Discovery 33(2), 293--324 (2019)

\bibitem[{Zhang et~al.(2010)Zhang, Yuan, Zhao, Fan, Zheng, and Wang}]{rdt_ml}
Zhang, X., Yuan, Q., Zhao, S., Fan, W., Zheng, W., Wang, Z.: {Multi-label
  Classification without the Multi-label Cost}.
\newblock In: Proceedings of the Society for Industrial and Applied Mathematics
  International Conference on Data Mining. pp. 778--789 (2010)

\bibitem[{{Zhang} and {Jung}(2019)}]{zhang2019GBDT-MO}
{Zhang}, Z., {Jung}, C.: {GBDT-MO: Gradient Boosted Decision Trees for Multiple
  Outputs}.
\newblock ArXiv preprints arXiv:1909.04373 [cs.CV] (2019)

\end{thebibliography}

\end{document}